\documentclass[conference]{IEEEtran}
\usepackage{cite}
\usepackage{amsmath,amssymb,amsfonts}
\usepackage[noend]{algorithmic}
\usepackage{algorithm}
\usepackage{graphicx}
\usepackage[table,xcdraw]{xcolor}
\usepackage{textcomp}
\usepackage{mwe}
\usepackage[normalem]{ulem}
\useunder{\uline}{\ul}{}
\usepackage{layouts}
\usepackage{comment}
\usepackage{multirow}
\usepackage{paralist}
\usepackage{url}
\usepackage{microtype}
\usepackage{calc}
\usepackage{booktabs}
\usepackage{xurl}
\usepackage{pifont}
\usepackage{mathrsfs}
\usepackage{subcaption}
\usepackage{hyperref}

\newcommand{\iconheight}{1.6ex}

\newcommand{\icon}[1]{%
  \raisebox{-0.2ex}{\includegraphics[height=\iconheight]{#1}}%
}

\DeclareRobustCommand{\frozen}{\icon{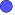}}
\DeclareRobustCommand{\learnable}{\icon{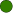}}
\DeclareRobustCommand{\mask}{\icon{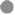}}
\DeclareRobustCommand{\masked}{\icon{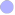}}
\DeclareRobustCommand{\maskfrozen}{\icon{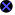}}

\def\BibTeX{{\rm B\kern-.05em{\sc i\kern-.025em b}\kern-.08em
    T\kern-.1667em\lower.7ex\hbox{E}\kern-.125emX}}
\begin{document}

\title{Don't Look Back in Anger: MAGIC Net for Streaming Continual Learning with Temporal Dependence}

\author{\IEEEauthorblockN{1\textsuperscript{st} Federico Giannini}
\IEEEauthorblockA{\textit{DEIB - Politecnico di Milano}, \\ federico.giannini@polimi.it,  \\ \href{https://orcid.org/0000-0002-4210-6271}{0000-0002-4210-6271}}
\and
\IEEEauthorblockN{2\textsuperscript{nd} Sandro D'andrea}
\IEEEauthorblockA{\textit{DEIB - Politecnico di Milano},\\
sandro.dandrea@mail.polimi.it}
\and
\IEEEauthorblockN{3\textsuperscript{rd} Emanuele Della Valle}
\IEEEauthorblockA{\textit{DEIB - Politecnico di Milano}, \\
emanuele.dellavalle@polimi.it,\\ \href{https://orcid.org/0000-0002-5176-5885}{0000-0002-5176-5885}}
}

\newcommand\blfootnote[1]{%
  \begingroup
  \renewcommand\thefootnote{}\footnotetext{#1}%
  \addtocounter{footnote}{-1}%
  \endgroup
}

\maketitle

\begin{abstract}
Concept drift, temporal dependence, and catastrophic forgetting represent major challenges when learning from data streams. While Streaming Machine Learning and Continual Learning (CL) address these issues separately, recent efforts in Streaming Continual Learning (SCL) aim to unify them. In this work, we introduce MAGIC Net, a novel SCL approach that integrates CL-inspired architectural strategies with recurrent neural networks to tame temporal dependence. MAGIC Net continuously learns, looks back at past knowledge by applying learnable masks over frozen weights, and expands its architecture when necessary. It performs all operations online, ensuring inference availability at all times. Experiments on synthetic and real-world streams show that it improves adaptation to new concepts, limits memory usage, and mitigates forgetting.
\end{abstract}


\blfootnote{\copyright 2024 IEEE.
This is the author’s accepted manuscript of the paper:
\mbox{Giannini, F.,} D'Andrea S., Della Valle, E. (2025). \textbf{Don't Look Back in Anger: MAGIC Net for Streaming Continual Learning with Temporal Dependence}. Proceedings of IEEE Big Data 2025, pp. 1396-1403.
The final authenticated version is available online at: \mbox{\url{https://doi.org/10.1109/BigData66926.2025.11401614}}}

\section{Introduction}
%\noindent The total text width of the document is: \the\textwidth 

%\noindent The total text width of the document is: \the\columnwidth
Changes in data distribution over time pose a crucial issue when applying machine learning models to data streams. Such changes are referred to as \textbf{concept drifts}~\cite{cit:concept_drift}.

Drifts can affect the distribution in two ways. The problem itself may be altered, so the new problem may \emph{contradict} previous ones. Let's think, for example, of product demand trend forecasting (i.e., growth or decline). Changes in consumer preferences, emerging technologies, or regulation may fundamentally alter how price, advertising, and availability influence demand. Conversely, the problem may be \emph{just expanded} by adding a new subproblem defined in a new input distribution, which does not contradict the previous and has no overlap. Moving back to the previous example, seasonal fluctuations may alter the overall distribution of product sales, keeping the relationship between factors like price and demand unchanged. 

Due to concept drifts, the model must \textbf{continuously learn} from new data and adapt to the new distributions (concepts). Models trained offline may, in fact, quickly become obsolete. Two research areas address drifts with different goals: Continual Learning (CL) and Streaming Machine Learning (SML).

CL~\cite{cit:cl} addresses an issue called \textbf{catastrophic forgetting}: when learning a new concept, one must not forget what it has learned from the past, since it is considered still valid~\cite{cit:catastrophic_forgetting}. The challenge lies in maintaining a balance between integrating novel concepts (plasticity) and preserving past information (stability), a problem known as the stability-plasticity dilemma\cite{cit:stability_plasticity}. This is particularly critical when new concepts do not contradict prior knowledge but complement it. CL assumes, thus, a scenario with only these types of drifts~\cite{cit:cl_survey_wang}.

SML~\cite{book_bifet} also addresses cases where the new problem contradicts the previous. It focuses on detecting drifts and quickly adapting to new concepts. It produces adaptive online learners and classifiers with the goal of performing as well as possible on the current distribution, ignoring the previous.  %Additionally, the model is updated each time the data stream generates a new data point.

Even if SML and CL have different objectives, they are both helpful in solving realistic, which may involve both types of drifts. When new problems may contradict the previous ones, the meaning of avoiding forgetting slightly changes. The solution cannot address all the concepts simultaneously, as they are mutually contradictory. Conversely, it should remember and isolate the knowledge associated with the previous concepts, although focusing on the new one. \textbf{Looking back at past concepts} may help solve future similar concepts or the recurrence of a previously observed concept.  

With these premises, the SML and CL communities started to cooperate. The new paradigm of Streaming Continual Learning (SCL) is emerging to blend the objectives of the two and solve real-world problems. SCL is a prominent research direction whose importance is highlighted by Gunasekara et al.~\cite{cit:scl_gunasekara}. Giannini et al.~\cite{cit:scl} deeply define it, and a dedicated special session at the ESANN 2025 conference~\cite{cit:scl_esann} consolidates it. 

Another critical problem emerging from the streaming research communities is the presence of \textbf{temporal dependence}~\cite{Ziffer22,cit:streaming_ts_zliobaite,cit:streaming_ts_read,cit:tenet}. Even if SML and CL usually neglect it, not considering it may cause inaccurate predictions. Data streams are, in fact, often associated with the Internet of Things, robotics, object detection, and satellite images. In these scenarios, the current observation strongly depends on the previous. This is also true for the previously presented example of product demand forecasting. Applying Time Series Analysis (TSA) is meaningful to exploit temporal patterns.
%Formally, given a data point $d_t$, there exists a lag $\tau$ such that \mbox{$\exists \tau \; P(a_t \mid b_{t-\tau}) \neq P(a_t)$}, where $a_t \in X_t \cup \{y_t\}$ and $b_{t-\tau} \in X_{t-\tau} \cup \{y_{t-\tau}\}$. This is especially critical when the current label $y_t$ depends on prior labels or features and looking at the previous data points when performing inference is crucial.

Following this direction, some recent works incorporate SCL and TSA. In particular, Continuous Progressive Neural Networks (cPNN)~\cite{cit:cpnn,cit:cpnn_extended} pioneers using CL strategies with Recurrent Neural Networks (RNN) to jointly deal with continuous learning, concept drifts, forgetting, and temporal dependence.

Despite the growing interest in SCL, existing approaches have primarily focused on isolated aspects of the problem, often neglecting the interplay between continuous learning, concept drift, temporal dependence, and forgetting. In the current literature, cPNN represents one of the few attempts to address these challenges simultaneously. However, there is still a need for novel solutions that can effectively integrate these aspects into a unified framework. For these reasons, we aim to investigate the following \textbf{research question}: \textit{how can a novel SCL solution be designed to effectively integrate continuous learning, concept drift, temporal dependence, and forgetting?}

This work explores a new SCL approach inspired by CL architectural strategies. Our \textbf{key contribution} is \textbf{MAGIC Net} (Masked, Adaptive, Growing, Intelligent, and Continuous Network). It is based on an RNN model trained continuously on accumulated mini-batches of data points. When a drift is detected by an external drift detector, it \emph{looks back at past knowledge} by learning real-valued masks in $(0,1)$ to be applied to the frozen weights learned during previous concepts. Additionally, it automatically decides if merely learning the masks is enough or if it needs to expand the network architecture by adding new weights. The great novelty here is that this process is performed online without splitting into separate offline phases. Unlike cPNN, it expands the architecture only when necessary. MAGIC Net can boost the adaptation to new concepts while reducing the memory impact. Storing prior masks in a persistent memory mitigates forgetting.

The rest of the paper is organized as follows. Section~\ref{sec:related_works} reviews related work. Section~\ref{sec:method} introduces MAGIC Net, our main contribution. Section~\ref{sec:experiments} details the experimental setup, and Section~\ref{sec:results} reports the results. Finally, Section~\ref{sec:conclusion} presents conclusions, limitations, and future work.

\section{Related Work}\label{sec:related_works}
A data stream can be logically defined as an unbounded sequence of data points $DS: d_0, d_1, ..., d_t, d_{t+1}, ...$ with $t \in \mathbb{N}$. Each point $d_t$ is a tuple $(X_t, y_t)$, where $X_t$ is the feature vector and $y_t$ the label, which becomes available only after predicting $\hat{y}_t$. For simplicity, the stream can be phisically defined as a sequence of $(X_t, y_{t-1})$, assuming that $y_{t-1}$ arrives after predicting $\hat{y}_{t-1}$ (given $X_{t-1}$) but before observing $X_t$. This setup enables \emph{prequential evaluation}~\cite{cit:prequential}, which at each step: \mbox{(1) updates} metrics with $\hat{y}_{t-1}$ vs. $y_{t-1}$, \mbox{(2) updates} the model with $(X_{t-1}, y_{t-1})$, and \mbox{(3) predicts} $\hat{y}_t$ from $X_t$. For instance, in environmental monitoring, the goal may be to predict if the temperature at time $t+1$ will exceed a threshold. The true label $y_t$ is revealed at $t{+}1$ with the new reading.

A \emph{concept drift}~\cite{cit:concept_drift} can be defined as an unpredictable change in the probability distribution $P(X,y)$, requiring the model to be updated. A \emph{concept} is the hidden process generating the data according to the distribution. Drifts can occur at different speeds: \emph{abrupt} drifts occur instantaneously, while \emph{gradual} and \emph{incremental} drifts unfold over time. Additionally, concepts may recur.

To learn continuously while taming temporal dependence, recent work follows the direction outlined by prior TSA studies, highlighting the forecasting potential of deep learning models~\cite{cit:makridakis_m5}. Neto et al.~\cite{cit:ilstm} propose a streaming application that accumulates data points generated by a data stream in fixed-size mini-batches. Whenever a mini-batch is full, it applies a sliding window to build sequences, and trains an RNN model for several epochs. This RNN model has multiple layers. Its many-to-many architecture produces inference only on mini-batches of data points. Subsequent work~\cite{cit:cpnn,cit:cpnn_extended,cit:tenet} elaborates on it and converges to a \textbf{Continuous RNN (cRNN)} version with a single layer to facilitate its application in streaming contexts. cRNN has a many-to-one architecture to optimize the inference whenever the data stream generates a new data point. The training procedure is still performed on mini-batches to have more accurate estimates of the loss function. cRNN can be applied to all types of RNN models.

Merely applying cRNN on data streams is not enough. Although it can learn continuously and tame temporal dependence, cRNN can suffer when the concept drift strongly changes the problem. In this case, the learning procedure may require several steps to converge to the minimum of the loss function associated with the new problem. Additionally, cRNN forgets the previous concept when it has learned the current one.

CL uses three main strategies to mitigate forgetting~\cite{cit:cl}. \emph{Replay} methods mix examples from past concepts with current ones, while \emph{Regularization} methods add terms to the loss. In realistic SCL scenarios with drifts affecting the learned decision boundary, these may fail to converge. Conversely, \emph{Architectural} methods handle one concept at a time, expand the architecture, share weights, apply transfer learning, isolate concept-specific weights, and freeze weights to maintain earlier knowledge.

CL assumes that, at each timestamp, the stream generates a large batch (called an experience) containing all the data points associated with the new concept. The model can process each experience without time constraints. To approach a more realistic streaming scenario, Online CL~\cite{cit:ocl_1} proposes an online version of CL that learns continuously from mini-batches containing tens of data points. Yet, it still focuses on avoiding forgetting in scenarios involving only non-contradicting drifts.

With these premises, \textbf{Continuous Progressive Neural Networks (cPNN)}~\cite{cit:cpnn,cit:cpnn_extended} represents a pioneering SCL solution. cPNN applies the Progressive Neural Networks (PNN)~\cite{cit:pnn} strategy on top of cRNN. cPNN architecture starts with a cRNN model (\emph{column}) trained continuously using mini-batches and runs in inference mode whenever the data stream generates a new $X_t$. After a drift is detected, it freezes the weights of the current column. It then adds a new column. For each item of an input sequence, the current column receives the concatenation between the feature vector representing that item and the associated hidden state of the previous column. This way the network can reuse the cumulated knowledge gained during the previous concepts and combine it with information coming from the current concept's data points. Freezing previous columns' weights avoids forgetting.

Regarding CL architectural strategies, \textbf{Piggyback~(PB)}~\cite{cit:piggyback} leverages a pre-trained backbone network. Whenever it receives an experience representing a new concept, it learns real-valued masks to apply to the frozen weights of the backbone network. During the forward pass, masks are binarized using a binarizer, allowing the selection of an optimal subnetwork. The backbone remains frozen to retain general knowledge. PB only learns masks for each experience. The masks are applied to the original weights via element-wise multiplication. PB strongly relies on the availability of a suitably pre-trained model.

\textbf{Compacting, Picking, and Growing (CPG)}~\cite{cit:cpg} follows a reverse approach, starting with a small network and expanding it progressively. It overcomes the need for pre-training and begins by training an initial network on the first experience. Afterward, it applies a gradual pruning process to pick only the relevant weights, denoted as $W^P_1$, that meet an accuracy goal. It then freezes these selected weights, ensuring they cannot be modified in subsequent training. The remaining weights, referred to as $W^E_1$, are left available for modifications during training on future tasks. The model learns a set of binary masks using the PB approach for each new experience $k$ (starting from the second). These masks are applied to the frozen weights $W^P_{k-1}$, while simultaneously training the remaining weights $W^E_{k-1}$. If the accuracy goal is not met at the end of the training, CPG resets the weights $W^E_{k-1}$ and expands them by adding new weights. The training then resumes, and the whole process is run iteratively until the accuracy goal is not met. Finally, CPG performs another round of gradual pruning to select the relevant weights $W^P_k$, while the remaining $W^E_k$ weights are left available for future experiences. CPG balances memory efficiency and adaptability, avoiding unnecessary model expansions. However, it can be applied only to classical CL scenarios that accumulate each concept's data points in a specific experience. CPG runs the phases (training, expanding, pruning) sequentially and cannot be run online.

Conversely, SML focuses on continuously learning from single data points while managing concept drifts. \textbf{Adaptive Random Forests (ARF)}~\cite{ARF} represent a streaming adaptive version of Random Forests, where each tree integrates concept drift detectors to adapt to drifts. The only solution to handle temporal dependence in SML is \textbf{Temporal Augmentation (TA)}~\cite{cit:streaming_ts_zliobaite}. Given a temporal augmentation order $o$, it expands the feature space of each data point by incorporating the previous $o$ labels, formally defined as \mbox{$X^{TA}_t = X_t \cup \{y_{t-1}, ..., y_{t-o}\}$}. However, SML ignores the forgetting problem since it aims to recover quickly after drifts.

\begin{figure*}[t]
\centering
\includegraphics{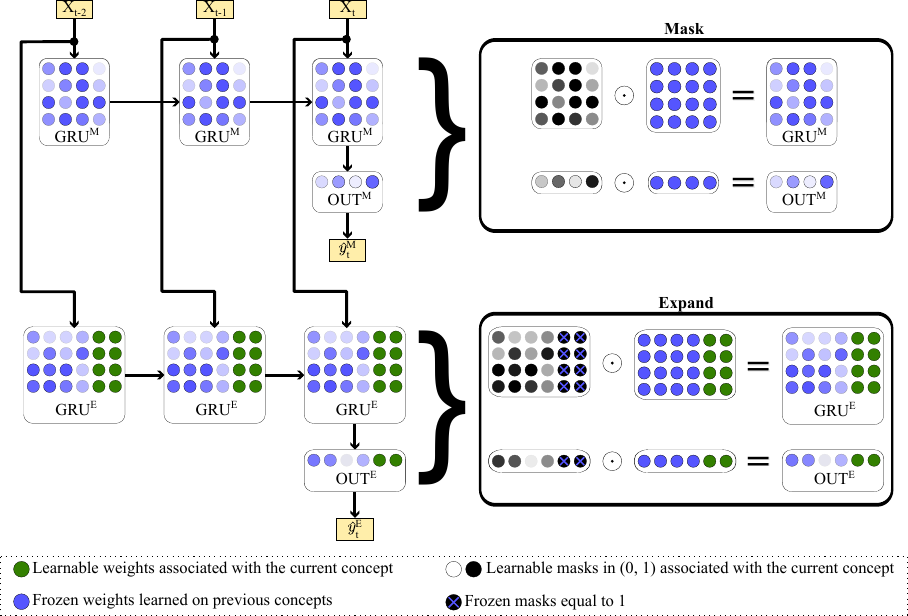}
\caption{Forward step on the feature vector $X_t$ during the ensemble phase.  For simplicity, only one \texttt{Mask} option is shown to represent both \texttt{MaskFineTune} and \texttt{MaskRandom}. They build weights (GRU$^M$, OUT$^M$) by multiplying the frozen cGRU weights by masks in $(0,1)$. Masks are obtained by applying a sigmoid to learnable values. \texttt{Expand} expands the GRU layer with new learnable weights. It applies learnable masks to the frozen cGRU weights, and frozen masks equal to 1 to the new learnable weights. It obtains GRU$^E$ and OUT$^E$. The opacity of the masked weights increases as the mask approaches 1.}\label{fig:magic_net}
\end{figure*}

\section{Proposed Method}\label{sec:method}
This work proposes \textbf{MAGIC Net} (Masked, Adaptive, Growing, Intelligent, and Continuous Network), a novel SCL approach that handles concept drifts, forgetting, and temporal dependence while learning continuously from a data stream. cPNN is the only attempt to solve this, but it has a major drawback: it always expands the architecture after a concept drift. The goal is to find an adaptive solution that can enhance concept drift adaptation by \textbf{looking back at the past knowledge} and can decide when expanding the architecture is needed to incorporate new knowledge. All the code and data used in this work are available for \textbf{reproducibility}.\footnote{\url{https://github.com/Sandrodand/MagicNet}}

MAGIC Net builds on cRNN (Section~\ref{sec:related_works}), which handles temporal dependence while learning continuously. This work uses the cGRU version, as it suits streaming scenarios requiring speed and simplicity. MAGIC Net relies on an external drift detector that alerts for detections, similar to how cPNN works.

It starts with a cGRU trained continuously on mini-batches, with all learnable weights. Fig.~\ref{fig:magic_net} shows the architecture after a drift detection. It freezes all the cGRU weights (\frozen) and builds a parallelizable ensemble containing the following options.
\begin{itemize}
    \item \texttt{MaskRandom} and \texttt{MaskFineTune} just learn masks to apply to the cGRU frozen weights. They learn real values, which are passed through a sigmoid function to obtain masks whose values are constrained within $(0,1)$ (\mask). Masks are then multiplied element-wise by the cGRU frozen weights (\frozen), producing the masked weight versions (\masked). While \texttt{MaskRandom} randomly initializes the values to which the sigmoid is applied, \texttt{MaskFineTune} starts from a copy of the previous concept's ones.
    \item \texttt{Expand} learns masks applied to the frozen cGRU weights and simultaneously expands the GRU hidden layer with new learnable weights to incorporate additional knowledge. 
    To achieve this, it creates a copy of the frozen cGRU and adds new weights to the GRU layer, increasing the hidden size by \texttt{expSize}. 
    The newly added weights are learnable (\learnable). Learnable masks within $(0,1)$ (\mask) are applied to the frozen cGRU weights (\frozen). Accordingly, frozen masks with values fixed to 1 (\maskfrozen) are applied to the newly added learnable weights (\learnable).

\end{itemize}

The usage of the ensemble enables MAGIC Net to determine whether it needs to expand the architecture or if applying masks to the previously learned weights is sufficient. By using both \texttt{MaskRandom} and \texttt{MaskFineTune}, it can decide if fine-tuning the previous masks is better than starting from random.

The ensemble runs in parallel for a specified number \texttt{numBatches} of training mini-batches following the detection. Whenever a new data point's feature vector $X_t$ is generated, MAGIC Net outputs the prediction of the currently best-performing option within the ensemble (considering the prequential evaluation on the data points following the detection). Given the window size \texttt{W}, the prediction is made considering the sequence $X_{t-W+1}, ..., X_t$. When the real label $y_t$ is available, it adds $(X_t,y_t)$ to the training mini-batch (as in cGRU training). When the mini-batch reaches a size \texttt{B}, it applies a sliding window to build sequences with size \texttt{W}. It then trains the masks of the three options and the new weights of the \texttt{Expand} option. To limit the computational overhead of the ensemble, after \texttt{numBatches} training mini-batches, the option with the best performance is kept, while the others are discarded.

The masks follow the training principle of PB and CPG. However, as already mentioned, MAGIC Net applies a sigmoid to bring them into a $(0,1)$ range, instead of simply binarizing them. This way, it increases the representational power of the masks. A binary mask, in fact, can just activate or deactivate a specific path (weight) within the network. By applying an element-wise multiplication between the weights and masks in $(0,1)$, MAGIC Net can consider portions of the weights without being limited to simply activating or deactivating them. Additionally, since the sigmoid is a differentiable non-linear function, it enables the exact calculation of the gradients of the actual mask values, facilitating and improving their training. This would not be possible with the non-differentiable PB Binarizer function. Sigmoid is applied to the learnable real values before element-wise multiplying them by the weights. 

%This is a method used to reduce the memory footprint of parameter tensors in neural networks by approximating them and using less precise formats compared to floating-point numbers. Usually, the floating-point weights are multiplied by a scale factor and then rounded to obtain integer representations of them. After this process, the masks are stored, while new ones are initialized for the next concept.

\section{Experimental Campaign}\label{sec:experiments}

%This section provides an overview of the setup of the experiments. Section~\ref{subsec:data_streams} explains the data streams' generation. Section~\ref{subsec:hypothesis} covers hypothesis formulation and the evaluation.

\subsection{Data Streams}\label{subsec:data_streams}
Although temporal dependence is a crucial problem, existing SML benchmarks lack the necessary temporal correlations. We, thus, use a benchmark incorporating temporal dependence.

\begin{equation} \label{eq:sine1_generalized} S_1: x_1 - \alpha - \beta \sin(\gamma x_2) = 0,\;S_2: x_1 - \alpha - \beta \sin(\gamma \pi x_2) = 0 \end{equation}

Following cPNN's evaluation~\cite{cit:cpnn_extended}, we use the \textbf{SineRW (SRW)} generator~\cite{cit:cpnn}, which produces 2D points ($x_1$, $x_2$) in $(0,1)$ via random walk and labels them using boundary functions. We extend it with two families of functions, $S_1$ and $S_2$~(\ref{eq:sine1_generalized}), representing \emph{simple} and \emph{complex} boundaries. Each function defines two binary classifiers: one assigns label 1 to points satisfying $\geq 0$, the other assigns 1 with condition $< 0$.

\begin{equation} \label{eq:sine_rw_mode} y_t' = MODE(y_t, y_{t-1}, y_{t-2}, y_{t-3}, y_{t-4}) \end{equation}

Next, we introduce a more complex temporal dependence. The new label at time $t$, represented as $ y'_{t}$, is computed using~(\ref{eq:sine_rw_mode}), where $y_t$ is the binary label assigned by $S_1$ or $S_2$. This calculation depends on both current and previous labels.

We also consider real data sources.
\textbf{AirQuality}\footnote{\url{https://data.seoul.go.kr/dataList/OA-15526/S/1/datasetView.do}} contains hourly measurements of pollutants (SO2, NO2, O3, CO, PM2.5, and PM10) collected from 25 monitoring stations in Seoul between 2017 and 2020, each collecting about 34k data points.

\textbf{PowerConsumption}~\cite{cit:power_consumption} comprises minute-by-minute electrical consumption records from a household in Socheuax from December 2006 to November 2010 (around 2M+ records). It includes active power, reactive power, voltage, energy sub-metering across different household areas, and inferred consumption for unmetered devices. Missing values (1.25\%) are interpolated linearly, and we apply a 15-minute tumbling window to compute moving averages~\cite{cit:power_consumption_deep_learning}.

\textbf{Weather}~\cite{cit:weather} originates from the Agricultural Research Service of the U.S. Department of Agriculture. It contains hourly hydrometeorological data from two stations in southwestern Idaho (2004-2014), each providing 96k+ records. Features include air temperature, relative humidity, wind speed, wind direction, and dew point temperature.

\begin{equation}
\label{eq:f1+}
\begin{split}
  F_{1+}: y(X_t) = 
  \begin{cases}
    1, & \text{if } v_t > v_{t-1} \\
    0, & \text{otherwise}
  \end{cases}
\end{split}
\end{equation}
\begin{equation}
\label{eq:f2+}
  \begin{aligned}
    F_{2+}: y(X_t) = 
    \begin{cases}
      1, & \text{if } v_t > Med(v_{t-k}, \ldots, v_{t-1}) \\
      0, & \text{otherwise}
    \end{cases}
  \end{aligned}
\end{equation}
\begin{equation}
\label{eq:f3+}
  F_{3+}: y(X_t) = 
  \begin{cases}
    1, & \text{if } v_t > Min(v_{t-k}, ..., v_{t-1})\\
    0, & \text{otherwise}
  \end{cases}
\end{equation}
\begin{equation}
\label{eq:f4+}
  F_{4+}: y(X_t) = 
  \begin{cases}
    1, & \text{if } \Delta_t > \Delta_{t-1}\\
    0, & \text{otherwise}
  \end{cases}
\end{equation}
\begin{equation}
\label{eq:f5+}
  F_{5+}: y(X_t) = 
  \begin{cases}
    1, & \text{if } \Delta_t > Med(\Delta_{t-k}, ..., \Delta_{t-1})\\
    0, & \text{otherwise}
  \end{cases}
\end{equation}

To construct binary classification problems, we standardize features and derive labels from a selected target variable $v_t$ (air temperature for Weather, NO2 for AirQuality, and active power for PowerConsumption). We define the five classification functions in (\ref{eq:f1+})-(\ref{eq:f5+}) (where $\Delta_t = v_t - v_{t-1}$). $k$ varies by dataset (10 for Weather, 5 for AirQuality, 20 for PowerConsumption). Five additional functions $F_{i-}$ reverse the labels of $F_{i+}$.  

Each data source gives rise to multiple data streams (\textbf{configurations}), which contain several concepts, each associated with a classification function. Each configuration randomly selects and combines the classification functions differently. Drifts are abrupt, and new concepts may contradict the previous. One concept per configuration may reintroduce a seen classification function.

\sloppy
SRWM considers 16 $S_1$ boundary functions ($\alpha \in {0,1}$, \mbox{$\beta \in {-1,1}$}, $\gamma \in [0.8,1.2]$) and 16 $S_2$ boundary functions ($\alpha=0.5$, $\beta \in [-0.25,-0.15]$, $\gamma \in [-2.2,-1.8]$). Each configuration contains eight concepts with 30k data points.

Weather's configurations concatenate data points from the two stations and split them into eight concepts. AirQuality considers eight concepts, each based on a different station's data. PowerConsumption considers five concepts.

\subsection{Experimental Setting}\label{subsec:hypothesis}
Our \textbf{hypotheses} are as follows: \textbf{H1)} \textit{MAGIC Net adapts more effectively to new concepts, outperforming SML models, cGRU, and cPNN.} \textbf{H2)} \textit{MAGIC Net outperforms the others when considering forgetting.} \textbf{H3)} \textit{MAGIC Net has a lower memory footprint than cPNN.}

We compare the following models for each configuration: ARF, ARF with TA (ARF$_T$), cGRU, cPNN, and MAGIC Net.

We apply two different types of evaluation. \textbf{Prequential evaluation}~\cite{cit:prequential} (which tests each data point's prediction before training on it) evaluates the online performance, focusing on single concepts. The performance at data point $d_t$ is calculated from the first data point following the last drift to $d_t$. We highlight two specific moments for each concept originated by the drift $j$: its start (\emph{start$_j$}) and its completion (\emph{end$_j$}). \emph{start$_j$} represents the first 50 learning mini-batches with size 128 ($50*128$ data points). We selected this moment since, as shown by Fig.~\ref{fig:prequential}, models start stabilizing after 50 mini-batches. It is useful to measure how the model reacts to the drift $j$. \emph{end$_j$} is the score computed on the last data point of the concept originated by the drift $j$. It is, thus, associated with all the data points belonging to the concept originated by the drift $j$. It is useful to measure how the model performs on the whole concept. We exclude the initial concept (the one before the first drift) since cGRU, cPNN, and MAGIC Net share identical architectures. We average the different start$_j$ and end$_j$ performance on all the concept drifts, producing start and end. The start performance represents the average performance after the first 50 mini-batches following the different drifts, end is the average performance after the end of the concepts.

\begin{equation} \label{eq:a_metric}
    AVG = \frac{\sum_{i=1}^{N} \sum_{j=1}^{i} R_{i,j}}{\frac{N \cdot (N+1)}{2}}
\end{equation}

\begin{equation} \label{eq:bwt_metric}
    BWT = \frac{\sum_{i=1}^{N} \sum_{j=1}^{i-1} (R_{i,j}-R_{j,j})}{\frac{N \cdot (N-1)}{2}}
\end{equation}

\textbf{CL evaluation} measures forgetting using model checkpoints saved at the end of each concept during prequential evaluation, where the last 2k points of each concept are discarded to form test sets. Each test set is input chronologically to the checkpoints, assuming $y_{t-1}$ is available at time $t$. This produces a matrix $R$, where $R_{i,j}$ is the score of the checkpoint after concept $i$ tested on concept $j$, using inference only. Average Metric (AVG) and Backward Transfer (BWT) are computed as in (\ref{eq:a_metric})(\ref{eq:bwt_metric}), where $N$ is the number of concepts~\cite{cit:cl}.

Since cPNN requires selecting the current column, when performing inference on a test set, one can run an ensemble containing all the columns. At timestamp $t$, the ensemble considers the prediction of the best-performing column on data points until $t-1$. After 500 data points, it only keeps the best-performing column. Similarly, MAGIC Net runs an ensemble containing all the saved masks and selects the best. 

Since the labels are not perfectly balanced, both evaluations compute \emph{Cohen's Kappa} score. 

After executing the preliminary experiments, \emph{hyperparameter} values are empirically selected as follows. Mini-batch size \mbox{\texttt{B}: 128}; Epochs number for each mini-batch: 10;  learning \mbox{rate: 0.01}; GRU hidden size: 50 for SRW, 25 for the real data sources; GRU sequence size \texttt{W}: 10 for SRW and AirQuality, 11 for Weather (to incorporate the temporal dependence of $F_2$), 48 for PowerConsumption (to consider 12 hours). We use the default settings for the SML models and refer to \texttt{W} to set the TA order $o$. For PowerConsumption, we consider $o$=20. MAGIC Net uses the following parameters: \texttt{expSize}: half of GRU hidden size; and \texttt{numBatches}: 30. %Additionally, to encourage the model to reuse past weights, the \texttt{Expand} options must outperform the \texttt{MaskRandom} ones by a factor of $105\%$. 

To focus on model performance independently of the specific drift detector employed, the campaign simulates multiple drift detectors with varying precision-recall. A true positive is any detection within 1k data points after a true concept drift (injected during stream construction). Given a target recall, the number of true positives is computed by rounding the product of recall and the number of true drifts. The total detections are then obtained by dividing true positives by the desired precision and rounding. Detection timestamps are randomly generated to match the desired performance. The considered precision-recall settings are: 100\%-100\%, 100\%-70\%, and 70\%-100\%.

\section{Results}\label{sec:results}
\begin{table*}[]
\centering
\begin{tabular}{ll|ll|ll|ll|ll|}
\multicolumn{1}{c}{\textbf{}} & \multicolumn{1}{c|}{\textbf{}} & \multicolumn{2}{c|}{\textbf{AirQuality}} & \multicolumn{2}{c|}{\textbf{PowerConsumption}} & \multicolumn{2}{c|}{\textbf{SRW}} & \multicolumn{2}{c|}{\textbf{Weather}} \\
\multicolumn{1}{c}{} & \multicolumn{1}{c|}{\textbf{}} & \multicolumn{1}{c}{\textbf{start}} & \multicolumn{1}{c|}{\textbf{end}} & \multicolumn{1}{c}{\textbf{start}} & \multicolumn{1}{c|}{\textbf{end}} & \multicolumn{1}{c}{\textbf{start}} & \multicolumn{1}{c|}{\textbf{end}} & \multicolumn{1}{c}{\textbf{start}} & \multicolumn{1}{c|}{\textbf{end}} \\ \hline
\multicolumn{1}{l|}{\multirow{5}{*}{\textbf{\begin{tabular}[c]{@{}l@{}}100\% prec\\ 100\% rec\end{tabular}}}} & ARF & 0.07±0.02 & 0.08±0.02 & 0.25±0.05 & 0.29±0.05 & 0.24±0.02 & 0.33±0.01 & 0.26±0.05 & 0.29±0.04 \\
\multicolumn{1}{l|}{} & ARF$_T$ & 0.21±0.05 & 0.26±0.04 & 0.22±0.06 & 0.28±0.06 & \textbf{0.73±0.00} & 0.73±0.00 & 0.41±0.06 & 0.44±0.06 \\
\multicolumn{1}{l|}{} & cGRU & 0.33±0.05 & 0.39±0.05 & 0.59±0.07 & 0.70±0.04 & 0.69±0.05 & \textbf{0.81±0.03} & 0.50±0.06 & 0.61±0.06 \\
\multicolumn{1}{l|}{} & cPNN & 0.35±0.03 & 0.35±0.03 & 0.66±0.02 & 0.71±0.02 & 0.61±0.05 & 0.68±0.04 & 0.52±0.05 & 0.56±0.05 \\
\multicolumn{1}{l|}{} & MAGIC & \textbf{0.42±0.03} & \textbf{0.47±0.03} & \textbf{0.68±0.04} & \textbf{0.77±0.02} & 0.71±0.04 & \textbf{0.82±0.03} & \textbf{0.56±0.04} & \textbf{0.67±0.03} \\ \hline
\multicolumn{1}{l|}{\multirow{5}{*}{\textbf{\begin{tabular}[c]{@{}l@{}}100\% prec\\ 70\% rec\end{tabular}}}} & ARF & 0.07±0.02 & 0.08±0.02 & 0.24±0.06 & 0.28±0.06 & 0.24±0.02 & 0.33±0.01 & 0.27±0.06 & 0.29±0.05 \\
\multicolumn{1}{l|}{} & ARF$_T$ & 0.20±0.05 & 0.26±0.05 & 0.21±0.06 & 0.27±0.07 & \textbf{0.73±0.00} & 0.73±0.00 & 0.42±0.07 & 0.45±0.06 \\
\multicolumn{1}{l|}{} & cGRU & 0.32±0.05 & 0.39±0.06 & 0.58±0.08 & 0.70±0.05 & 0.69±0.05 & \textbf{0.81±0.03} & 0.51±0.07 & 0.61±0.06 \\
\multicolumn{1}{l|}{} & cPNN & 0.34±0.03 & 0.35±0.03 & 0.65±0.05 & 0.70±0.03 & 0.63±0.04 & 0.70±0.04 & \textbf{0.53±0.05} & 0.59±0.05 \\
\multicolumn{1}{l|}{} & MAGIC & \textbf{0.40±0.03} & \textbf{0.46±0.03} & \textbf{0.67±0.05} & \textbf{0.76±0.03} & 0.70±0.05 & \textbf{0.81±0.04} & \textbf{0.54±0.06} & \textbf{0.66±0.05} \\ \hline
\multicolumn{1}{l|}{\multirow{5}{*}{\textbf{\begin{tabular}[c]{@{}l@{}}70\% prec\\ 100\% rec\end{tabular}}}} & ARF & 0.07±0.02 & 0.08±0.02 & 0.25±0.05 & 0.29±0.05 & 0.24±0.02 & 0.33±0.01 & 0.26±0.05 & 0.29±0.04 \\
\multicolumn{1}{l|}{} & ARF$_T$ & 0.21±0.05 & 0.26±0.04 & 0.22±0.06 & 0.28±0.06 & \textbf{0.73±0.00} & 0.73±0.00 & 0.41±0.06 & 0.44±0.06 \\
\multicolumn{1}{l|}{} & cGRU & 0.33±0.05 & 0.39±0.05 & 0.59±0.07 & 0.70±0.04 & 0.69±0.05 & \textbf{0.81±0.03} & 0.50±0.06 & 0.61±0.06 \\
\multicolumn{1}{l|}{} & cPNN & 0.34±0.02 & 0.34±0.03 & 0.65±0.03 & 0.70±0.02 & 0.59±0.07 & 0.66±0.06 & 0.50±0.05 & 0.53±0.05 \\
\multicolumn{1}{l|}{} & MAGIC & \textbf{0.43±0.03} & \textbf{0.48±0.02} & \textbf{0.67±0.04} & \textbf{0.77±0.02} & 0.69±0.04 & \textbf{0.81±0.03} & \textbf{0.55±0.05} & \textbf{0.66±0.03} \\ \hline
\end{tabular}
\caption{Results of the \textbf{prequential evaluation} with \textbf{Cohen's Kappa}. For each data source and detector, the mean and standard deviation over 50 configurations are shown. Bold indicates best models. ARF$_T$ and cGRU excel on SRW. MAGIC Net leads on real data, with cPNN slightly worse at start on PowerConsumption and Weather. H1 is supported on real data.}
\label{table:prequential_kappa}
\end{table*}

\begin{figure*}[!]
    \centering
    % First row
    \begin{subfigure}{0.32\textwidth}
        \centering
        \includegraphics[]{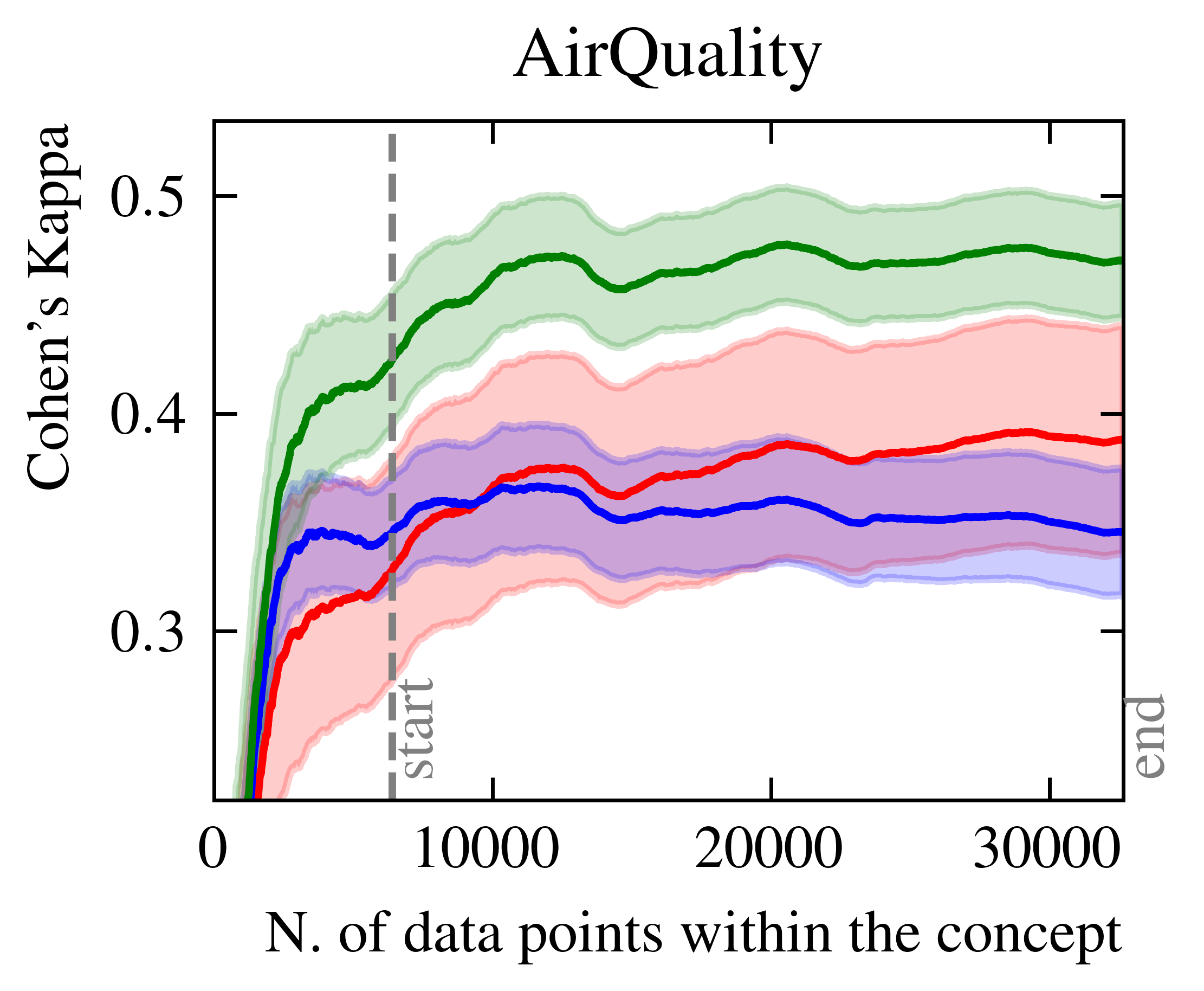}
        \phantomcaption{}
        \label{fig:performance_averaged_100_100_aq}
    \end{subfigure}
    \hfill
    \begin{subfigure}{0.32\textwidth}
        \centering
        \includegraphics[]{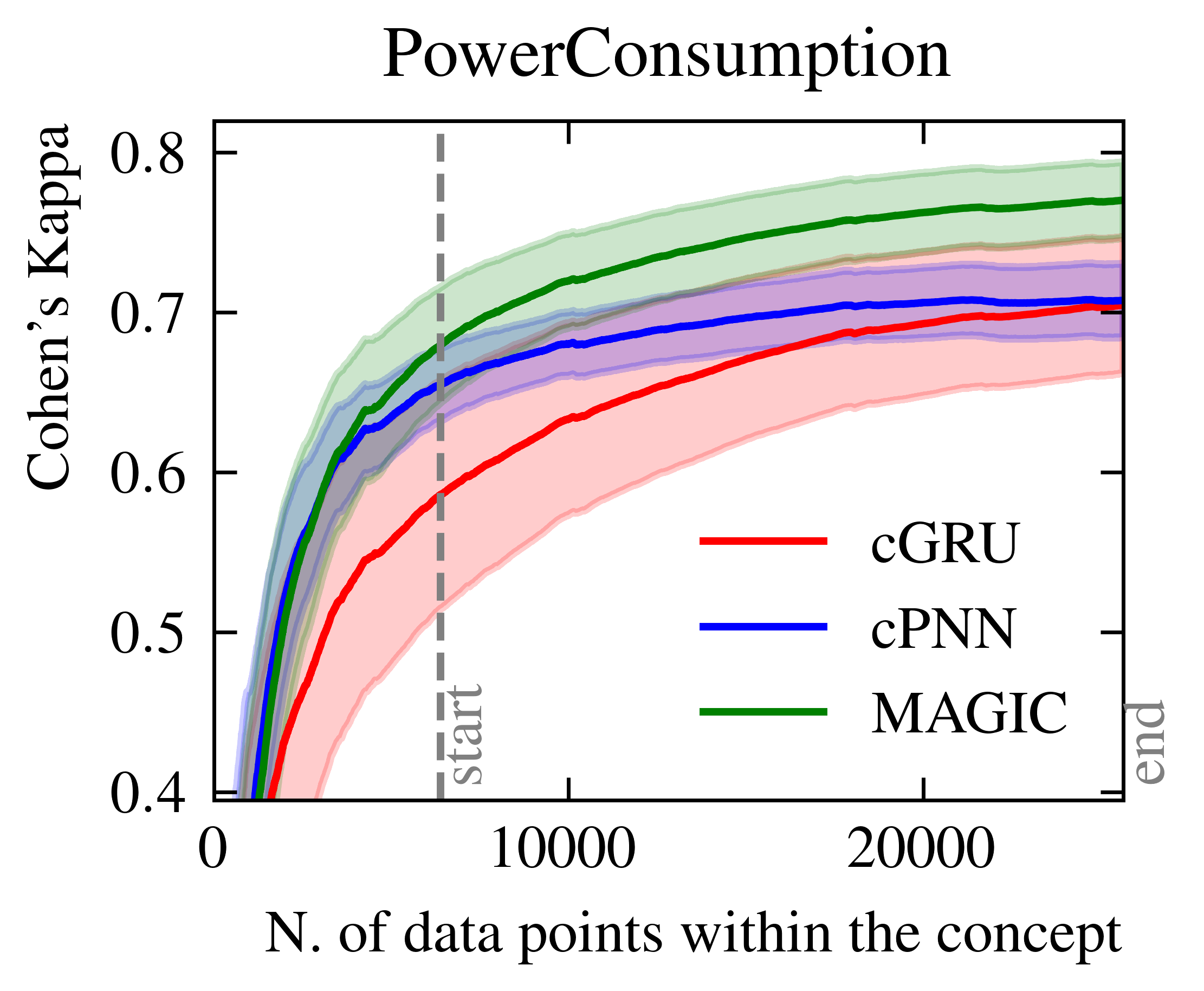}
        \phantomcaption{}
        \label{fig:performance_averaged_100_100_pc}
    \end{subfigure}
    \hfill
    \begin{subfigure}{0.32\textwidth}
        \centering
        \includegraphics[]{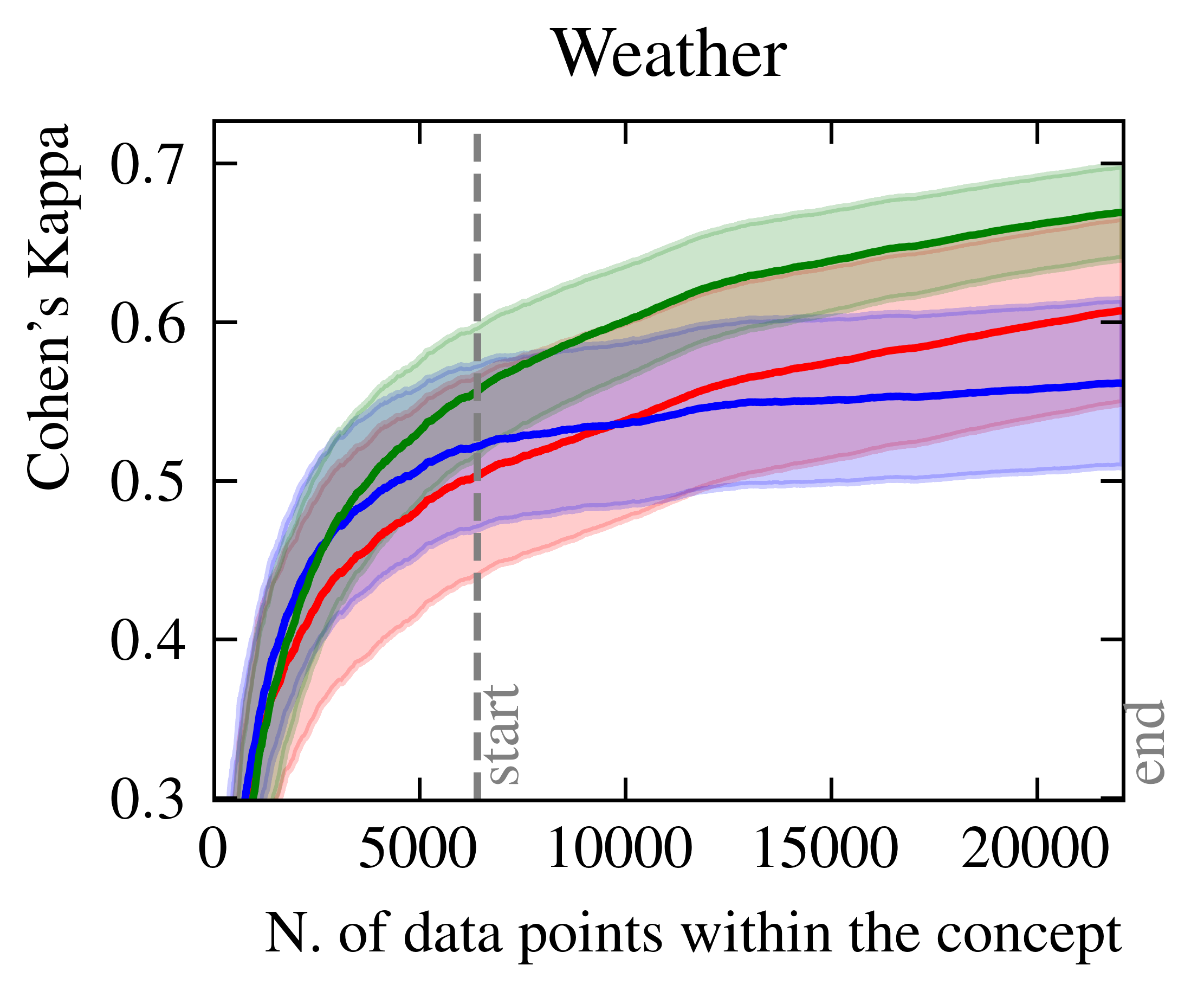}
        \phantomcaption{}
        \label{fig:performance_averaged_100_100_w}
    \end{subfigure}

    \caption{Average Cohen's Kappa over time on 50 configurations of real data, using a concept drift detector with 100\% precision and recall. Scores are first averaged per concept, then across configurations. MAGIC Net outperforms others across the stream. On PowerConsumption and Weather, cPNN starts slightly worse. MAGIC Net is clearly best at concept ends.}
    \label{fig:prequential}
\end{figure*}

\begin{figure*}[t]
    \centering
    % First row
    \begin{subfigure}{0.32\textwidth}
        \centering
        \includegraphics[]{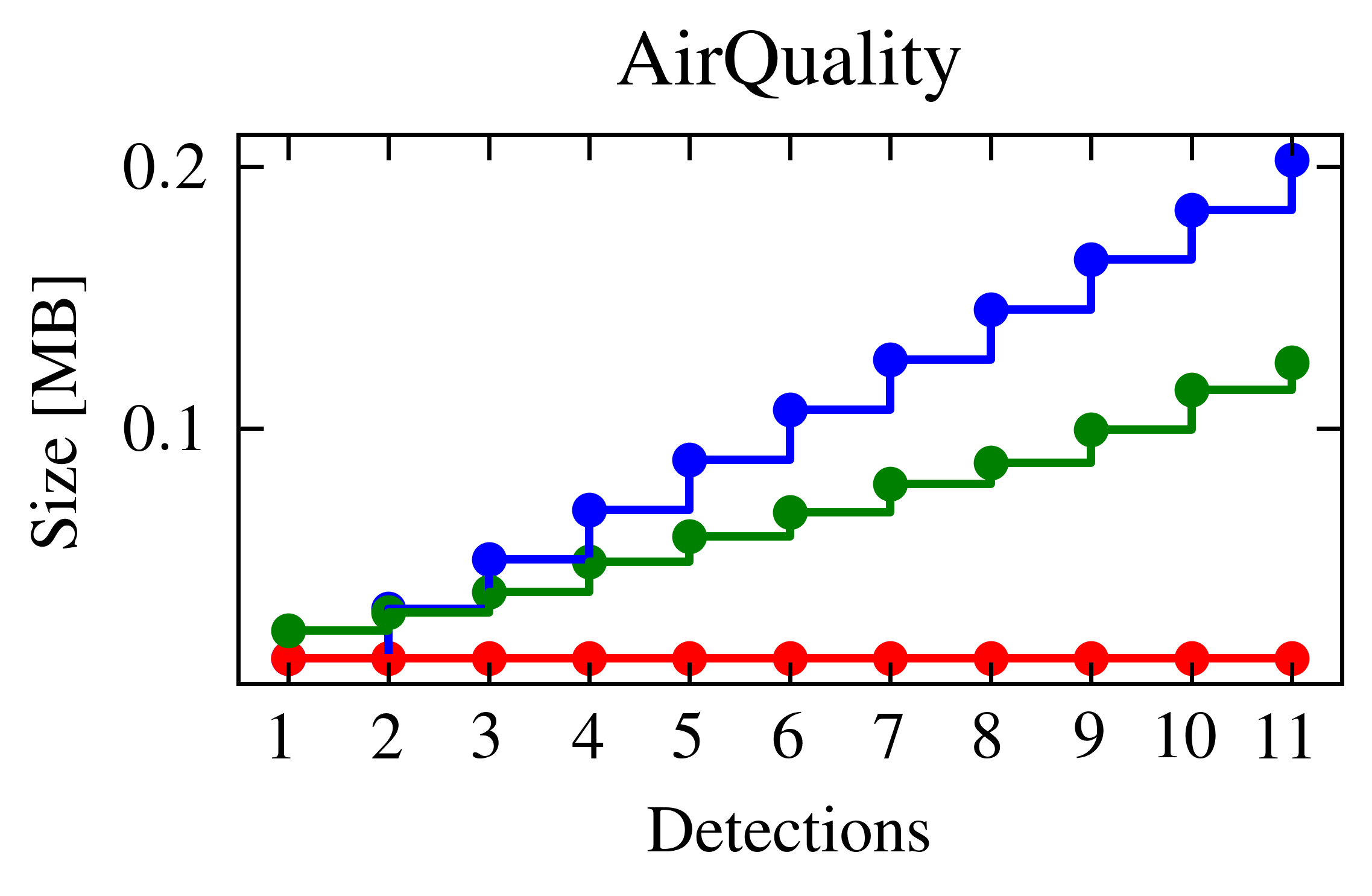}
        \phantomcaption{}
        \label{fig:memory_aq}
    \end{subfigure}
    \hfill
    \begin{subfigure}{0.32\textwidth}
        \centering
        \includegraphics[]{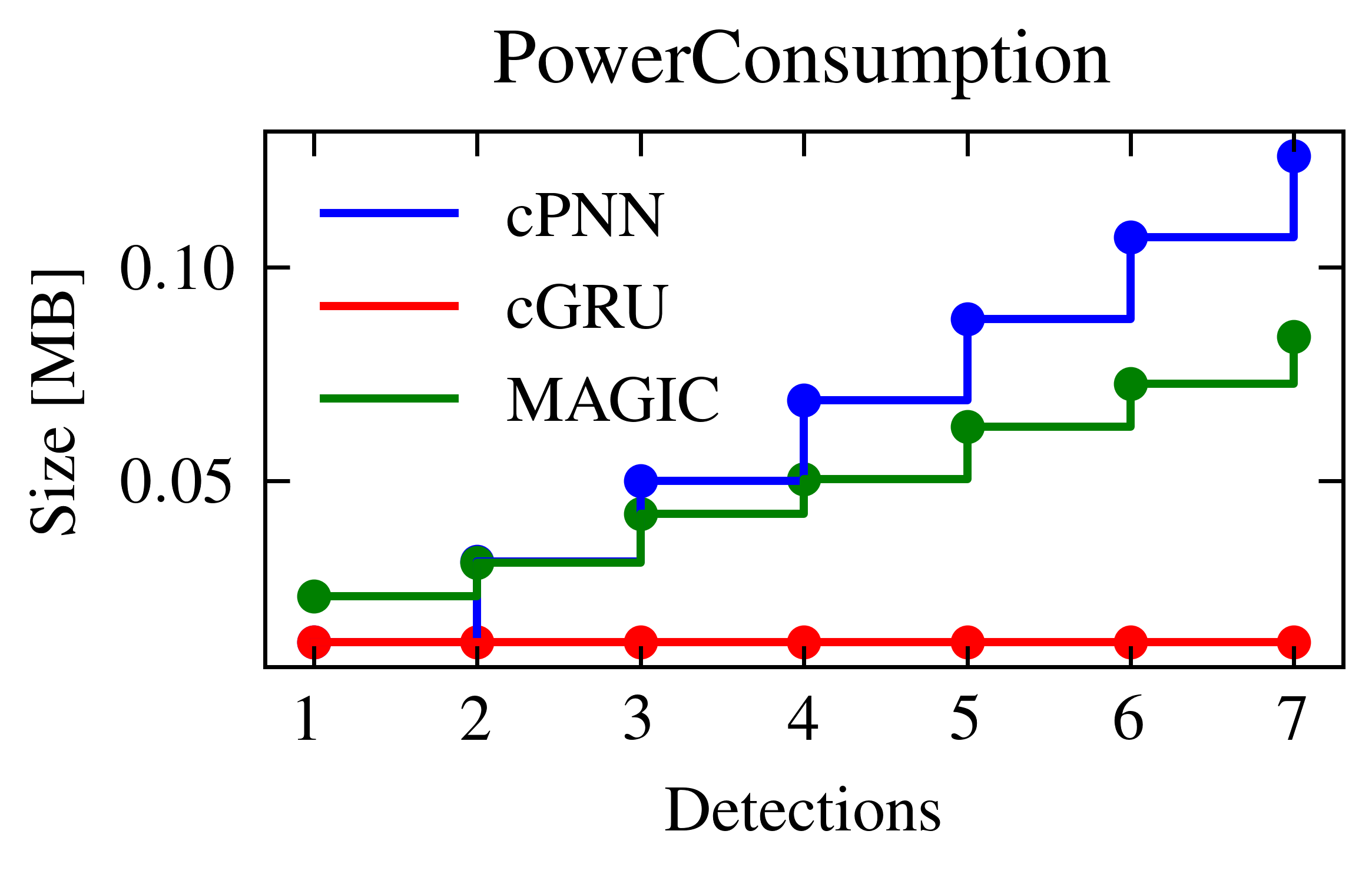}
        \phantomcaption{}
        \label{fig:memory_pc}
    \end{subfigure}
    \hfill
    \begin{subfigure}{0.32\textwidth}
        \centering
        \includegraphics[]{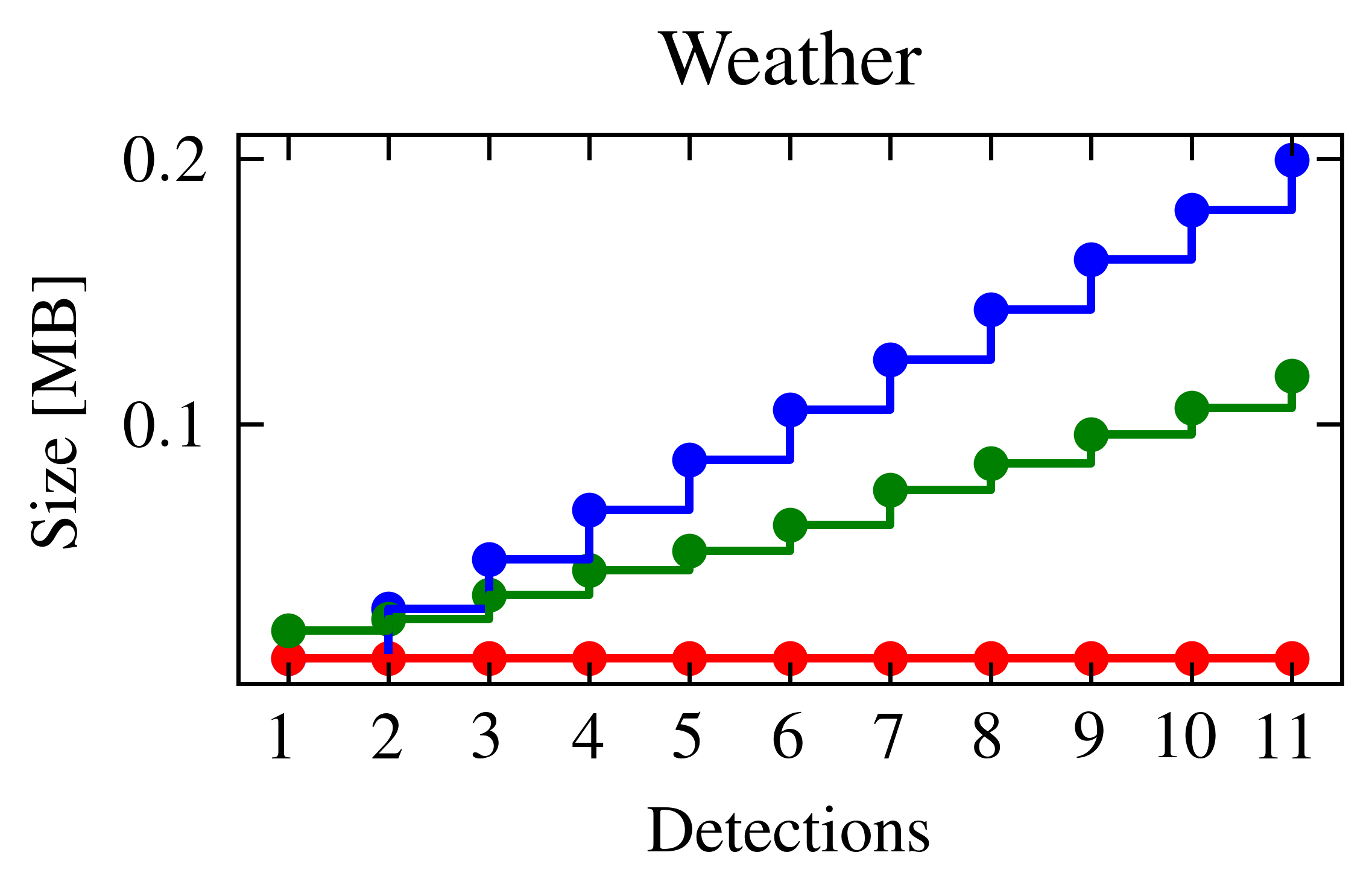}
        \phantomcaption{}
        \label{fig:memory_w}
    \end{subfigure}

    \caption{Average sizes in MB on the 50 configurations of the real data sources over the detections with 70\% precision and 100\% recall. MAGIC Net reduces the size as new detections arise by deciding when to expand the architecture. H3 is proven.}
    \label{fig:memory}
\end{figure*}

\begin{table*}[]
\centering
\begin{tabular}{ll|ll|ll|ll|ll|}
\multicolumn{1}{c}{\textbf{}} & \multicolumn{1}{c|}{\textbf{}} & \multicolumn{2}{c|}{\textbf{AirQuality}} & \multicolumn{2}{c|}{\textbf{PowerConsumption}} & \multicolumn{2}{c|}{\textbf{SRW}} & \multicolumn{2}{c|}{\textbf{Weather}} \\
\multicolumn{1}{c}{} & \multicolumn{1}{c|}{\textbf{}} & \multicolumn{1}{c}{\textbf{AVG}} & \multicolumn{1}{c|}{\textbf{BWT}} & \multicolumn{1}{c}{\textbf{AVG}} & \multicolumn{1}{c|}{\textbf{BWT}} & \multicolumn{1}{c}{\textbf{AVG}} & \multicolumn{1}{c|}{\textbf{BWT}} & \multicolumn{1}{c}{\textbf{AVG}} & \multicolumn{1}{c|}{\textbf{BWT}} \\ \hline
\multicolumn{1}{l|}{\multirow{5}{*}{\textbf{\begin{tabular}[c]{@{}l@{}}100\% prec\\ 100\% rec\end{tabular}}}} & ARF & 0.01±0.01 & -0.07±0.02 & 0.07±0.02 & -0.35±0.08 & 0.09±0.02 & -0.36±0.03 & 0.04±0.01 & -0.18±0.04 \\
\multicolumn{1}{l|}{} & ARF$_T$ & 0.14±0.04 & -0.19±0.05 & 0.09±0.02 & -0.31±0.12 & \textbf{0.73±0.00} & -0.00±0.00 & 0.20±0.05 & -0.24±0.06 \\
\multicolumn{1}{l|}{} & cGRU & 0.06±0.02 & -0.48±0.06 & 0.18±0.05 & -0.86±0.08 & 0.20±0.05 & -0.85±0.07 & 0.11±0.02 & -0.68±0.06 \\
\multicolumn{1}{l|}{} & cPNN & 0.31±0.05 & -0.09±0.05 & 0.51±0.09 & -0.32±0.15 & 0.59±0.10 & -0.21±0.12 & 0.42±0.09 & -0.19±0.09 \\
\multicolumn{1}{l|}{} & MAGIC & \textbf{0.41±0.06} & -0.11±0.05 & \textbf{0.55±0.10} & -0.35±0.15 & 0.64±0.11 & -0.26±0.14 & \textbf{0.49±0.07} & -0.24±0.08 \\ \hline
\multicolumn{1}{l|}{\multirow{5}{*}{\textbf{\begin{tabular}[c]{@{}l@{}}100\% prec\\ 70\% rec\end{tabular}}}} & ARF & 0.01±0.01 & -0.07±0.02 & 0.07±0.02 & -0.35±0.08 & 0.09±0.02 & -0.36±0.03 & 0.04±0.01 & -0.18±0.04 \\
\multicolumn{1}{l|}{} & ARF$_T$ & 0.14±0.04 & -0.19±0.05 & 0.09±0.02 & -0.31±0.12 & \textbf{0.73±0.00} & -0.00±0.00 & 0.20±0.05 & -0.24±0.06 \\
\multicolumn{1}{l|}{} & cGRU & 0.06±0.02 & -0.48±0.06 & 0.18±0.05 & -0.86±0.08 & 0.20±0.05 & -0.85±0.07 & 0.11±0.02 & -0.68±0.06 \\
\multicolumn{1}{l|}{} & cPNN & 0.28±0.06 & -0.13±0.07 & 0.47±0.11 & -0.39±0.16 & 0.55±0.09 & -0.29±0.13 & 0.36±0.07 & -0.29±0.10 \\
\multicolumn{1}{l|}{} & MAGIC & \textbf{0.36±0.06} & -0.16±0.07 & \textbf{0.50±0.09} & -0.42±0.13 & 0.59±0.09 & -0.33±0.13 & \textbf{0.42±0.07} & -0.30±0.09 \\ \hline
\multicolumn{1}{l|}{\multirow{5}{*}{\textbf{\begin{tabular}[c]{@{}l@{}}70\% prec\\ 100\% rec\end{tabular}}}} & ARF & 0.01±0.01 & -0.07±0.02 & 0.07±0.02 & -0.35±0.08 & 0.09±0.02 & -0.36±0.03 & 0.04±0.01 & -0.18±0.04 \\
\multicolumn{1}{l|}{} & ARF$_T$ & 0.14±0.04 & -0.19±0.05 & 0.09±0.02 & -0.31±0.12 & \textbf{0.73±0.00} & -0.00±0.00 & 0.20±0.05 & -0.24±0.06 \\
\multicolumn{1}{l|}{} & cGRU & 0.06±0.02 & -0.48±0.06 & 0.18±0.05 & -0.86±0.08 & 0.20±0.05 & -0.85±0.07 & 0.11±0.02 & -0.68±0.06 \\
\multicolumn{1}{l|}{} & cPNN & 0.34±0.05 & -0.05±0.05 & \textbf{0.60±0.08} & -0.18±0.13 & 0.65±0.07 & -0.10±0.08 & 0.46±0.07 & -0.10±0.07 \\
\multicolumn{1}{l|}{} & MAGIC & \textbf{0.44±0.06} & -0.10±0.07 & \textbf{0.61±0.09} & -0.28±0.13 & \textbf{0.72±0.07} & -0.16±0.10 & \textbf{0.53±0.06} & -0.19±0.07 \\ \hline
\end{tabular}
\caption{Results of the \textbf{CL evaluation} considering the \textbf{Cohen's Kappa}. SML models and cGRU suffer from forgetting, while cPNN and MAGIC Net are robust. MAGIC Net is the best-performing model. H2 is almost always verified.}
\label{table:cl}
\end{table*}

Table~\ref{table:prequential_kappa} shows the results of the prequential evaluation considering the values of Cohen's Kappa at start and end. Similar results were obtained considering balanced accuracy, which, due to space issues, are not reported. The table reports the mean and standard deviation over the 50 configurations for each data source and detector scenario. Drifts before the the first detection are ignored. The statistically best-performing models are highlighted in bold. To assess whether one model outperforms another, we conduct a statistical hypothesis test with $\alpha=0.05$, considering the average performance over the different configurations. We first perform a Shapiro-Wilk test to assess the normality of the differences between the two score lists. If the null hypothesis cannot be rejected, we proceed with a paired t-test. Otherwise, we use the Wilcoxon signed-rank test. One-sided tests are conducted.

When considering the synthetic data streams (SRW),  adding the TA on ARF (namely ARF$_T$) produces good performance. The classic ARF cannot learn since it does not consider temporal dependence. As a result, ARF$_T$ is always the best-performing model on the start performance. This result can be explained by the nature of the data stream. In SRW, the percentage of instances where the current label matches the previous one is  86.74\%. Since ARF$_T$ is inputted with the previous real labels, it quickly learns to predict the current label as the previous. The start and end values are equal and have a standard deviation of zero. Additionally, the temporal pattern never changes over the data stream since all the classification functions simply compute the mode. What ARF$_T$ learns during the first concept remains valid over the entire stream.

In this SRW simple problem, adding a strategy to the base cGRU learner is not helpful since cGRU performs well. Applying cPNN reduces the performance, while applying MAGIC Net keeps it more or less unchanged. At the end of the concepts, MAGIC Net and cGRU can outperform ARF$_T$.

The situation changes with real data sources. ARF$_T$ struggles to learn reliably and consistently underperforms compared to cGRU-based models, as it cannot capture complex temporal dependencies and depends too much on previous labels. The classification functions introduce frequent label changes, making the task complex and dependent on recognizing temporal patterns. The percentage of instances where the current label matches the previous drops to 63.08\% for AirQuality, 51.34\% for PowerConsumption, and 71.34\% for Weather.

Looking at AirQuality, MAGIC Net clearly outperforms all models for both start and end metrics across all detector scenarios. In this case, cPNN can outperform cGRU only during the initial data points of the concepts, but is later surpassed by it. Regarding PowerConsumption, MAGIC Net adapts to new concepts slightly better than cPNN (its start performance is marginally better with statistical evidence). At the end of the concepts, MAGIC Net significantly outperforms the others, with cGRU reaching cPNN's performance. These findings are consistent across all detector scenarios. For Weather, the results in the 100\%-100\% and 70\%-100\% recall-precision detector scenarios are similar to PowerConsumption. Additionally, cGRU outperforms cPNN at the end of the concepts. When recall is reduced to 70\%, cPNN and MAGIC Net perform equally at the start, but MAGIC Net outperforms the others at the end, with cGRU approaching cPNN's performance. \textbf{H1} is thus verified for the real data sources.

These findings are better visualized in Fig.~\ref{fig:prequential}, which shows the evolution of Cohen's Kappa score over time for the concepts of real data sources with 100\% recall and precision. For each configuration, it averages the scores of the concepts from the first drift onward and represents the distribution spread by plotting the standard deviation over time.

MAGIC Net can recognize false positive detections through its ensemble mechanism, which includes the \texttt{MaskFineTune} option for fine-tuning previous concept masks. In such cases, not expanding is usually the best-performing choice. This is supported by the average number of expansions: under both 100\%-100\% and 70\%-100\% precision-recall settings, expansion counts are similar. On AirQuality, MAGIC Net expands on average 3.56 vs. 3.42 times; on PowerConsumption, 2.16 vs. 2.36; on Weather, 3.16 vs. 3.28. Some true drifts may not require expansion, and reusing prior knowledge may be sufficient. MAGIC Net expands in 50\% of true drifts. In contrast, cPNN expands on every detection, making the number of expansions equal to the number of detections. This leads to higher memory usage, which increases inversely with precision. MAGIC Net limits this by expanding only when needed. Fig.\ref{fig:memory} illustrates the average model sizes (MB) over 50 configurations with 70\% precision and 100\% recall. These results verify \textbf{H3}.

Table~\ref{table:cl} reports CL evaluation results based on Cohen's Kappa. ARF, ARF$_T$, and cGRU suffer from forgetting, as they focus only on the current concept. ARF$_T$ performs well only on SRW, where the temporal pattern remains fixed. MAGIC Net achieves the highest AVG score across all real data sources and detectors. Its BWT is slightly worse than cPNN's due to its higher average Kappa. \textbf{H2} is consistently verified.

%The average global execution time is around 96 minutes on an AWS t3.2xlarge instance. For each data source, we summed the prequential and CL evaluations across three detector scenarios and five models for each of its 50 configurations, then averaged them. Then, we averaged across the four data sources.

\section{Conclusion}\label{sec:conclusion}
This work addressed the challenges of learning from real-world data streams by proposing MAGIC Net, a model that combines SCL and CPG principles. It looks back at the past knowledge by applying real-valued masks to frozen weights from past concepts, and expands its architecture when needed. Built on a continuously adapted RNN model to handle temporal dependence, MAGIC Net outperformed state-of-the-art methods in prequential evaluation on real-data scenarios. It limited memory usage by expanding selectively and proved robust to forgetting thanks to persistent mask storage.

Although it was evaluated with GRUs for their efficiency, MAGIC Net is modular and compatible with any sequence model. Moreover, even if it reduces memory usage, it still lacks architectural compaction. Adapting CPG pruning could be a promising direction.
From preliminary results, it appears that masks act as a regularizer. However, SCL solutions are sensitive to hyperparameters, and a systematic study beyond our empirical approach is needed. Future investigation may also explore other drift types, such as incremental or gradual. We leave to an extended version of this work a theoretical analysis of the semantics associated with sigmoid masks, as well as an ablation study highlighting the contribution of each component. Furthermore, we plan to employ real drift detectors to evaluate the approach in realistic scenarios.

\bibliographystyle{splncs04}
\bibliography{bibliography}

\end{document}